\documentclass{article}

\usepackage{arxiv}

\usepackage[utf8]{inputenc} % allow utf-8 input
\usepackage[T1]{fontenc}    % use 8-bit T1 fonts
\usepackage{hyperref}       % hyperlinks
\usepackage{url}            % simple URL typesetting
\usepackage{booktabs}       % professional-quality tables
\usepackage{amsfonts}       % blackboard math symbols
\usepackage{nicefrac}       % compact symbols for 1/2, etc.
\usepackage{microtype}      % microtypography
\usepackage{lipsum}
\usepackage{graphicx}
\usepackage{multirow}
\usepackage{floatrow}
\newfloatcommand{capbtabbox}{table}[][\FBwidth]
\graphicspath{ {./images/} }

\usepackage{multirow}

\usepackage{makecell}
\usepackage{xcolor}
\usepackage{colortbl}
\usepackage{url}
\usepackage{listings}
\usepackage{enumitem}

\usepackage[font=scriptsize]{subfig}

\definecolor{codegreen}{rgb}{0,0.5,0}
\definecolor{codeblue}{rgb}{0.25,0.5,0.5}
\definecolor{codegray}{rgb}{0.6,0.6,0.6}
\usepackage{color}

\definecolor{dkgreen}{rgb}{0,0.6,0}
\definecolor{gray}{rgb}{0.5,0.5,0.5}
\definecolor{mauve}{rgb}{0.58,0,0.82}

\usepackage{bm}

\usepackage{authblk}

\begin{document}

\title{PingAn-VCGroup's Solution for ICDAR 2021 Competition on Scientific Literature Parsing Task B: Table Recognition to HTML}

\author[1]{Jiaquan Ye}
\author[1]{Xianbiao Qi}
\author[1]{Yelin He}
\author[1]{Yihao Chen}
\author[1]{Dengyi Gu}
\author[2]{Peng Gao}
\author[1]{Rong Xiao}
\affil[1]{Visual Computing Group, Ping An Property \& Casualty Insurance Company}
\affil[2]{Ping An Technology Company}

%\author{
% Jiaquan Ye
%}
%\author[xxx]{aaa}
%\address{1. Visual Computing Group,
%  Ping An Property \& Casualty Insurance Company.\\
% 2. Ping An Technology Company.}
\thanks{Xianbiao Qi is the corresponding author. If you have any questions or concerns about the implementation details, please do not hesitate to contact jiaquanye@qq.com or qixianbiao@gmail.com.}

\maketitle
\begin{abstract}
This paper presents our solution for ICDAR 2021 competition on scientific literature parsing task B: table recognition to HTML. In our method, we divide the table content recognition task into four sub-tasks: table structure recognition, text line detection, text line recognition, and box assignment. Our table structure recognition algorithm is customized based on MASTER~\cite{lu2019master}, a robust image text recognition algorithm. PSENet~\cite{wang2019shape} is used to detect each text line in the table image. For text line recognition, our model is also built on MASTER. 
Finally, in the box assignment phase, we associated the text boxes detected by PSENet with the structure item reconstructed by table structure prediction, and fill the recognized content of the text line into the corresponding item. Our proposed method achieves a 96.84\% TEDS score on 9,115 validation samples in the development phase, and a 96.32\% TEDS score on 9,064 samples in the final evaluation phase. 
\end{abstract}

% keywords can be removed

\section{Introduction}
The ICDAR 2021 competition on scientific literature parsing task B is to reconstruct the table image into an HTML code. In this competition, PubTabNet dataset (v2.0.0)~\cite{zhong2019image} is provided as the official evaluation data, and Tree-Edit-Distance-based similarity (TEDS) metric is used for evaluation. The PubTabNet data set consists of 500,777 training samples, 9,115 validation samples, 9,138 samples for the development stage, and 9,064 samples for the final evaluation stage. For the training and validation data, the ground truth HTML codes and the position of non-empty table cells are provided to the participants. Participants of this competition need to develop a model that can convert images of tabular data into the corresponding HTML code, which should correctly represent the structure of the table and the content of each cell. The labels of samples for the development and the final evaluation stages are preserved by the organizers.

We divide this task into four sub-tasks: table structure recognition, text line detection, text line recognition, and box assignment. And several tricks are tried to improve the model. The details of each sub-task will be discussed in the following section.

%The rest of the paper is organized as follows. Firstly, we will introduce the methods of this competition in section 2. Some experiment results will be presented in section 3. And finally, section 4 concludes the paper. 

\section{Method}
\label{sec:headings}
In this section, we will present these four sub-tasks in order.

\begin{figure}[t] % picture
    \centering
    \includegraphics[width=0.85\textwidth]{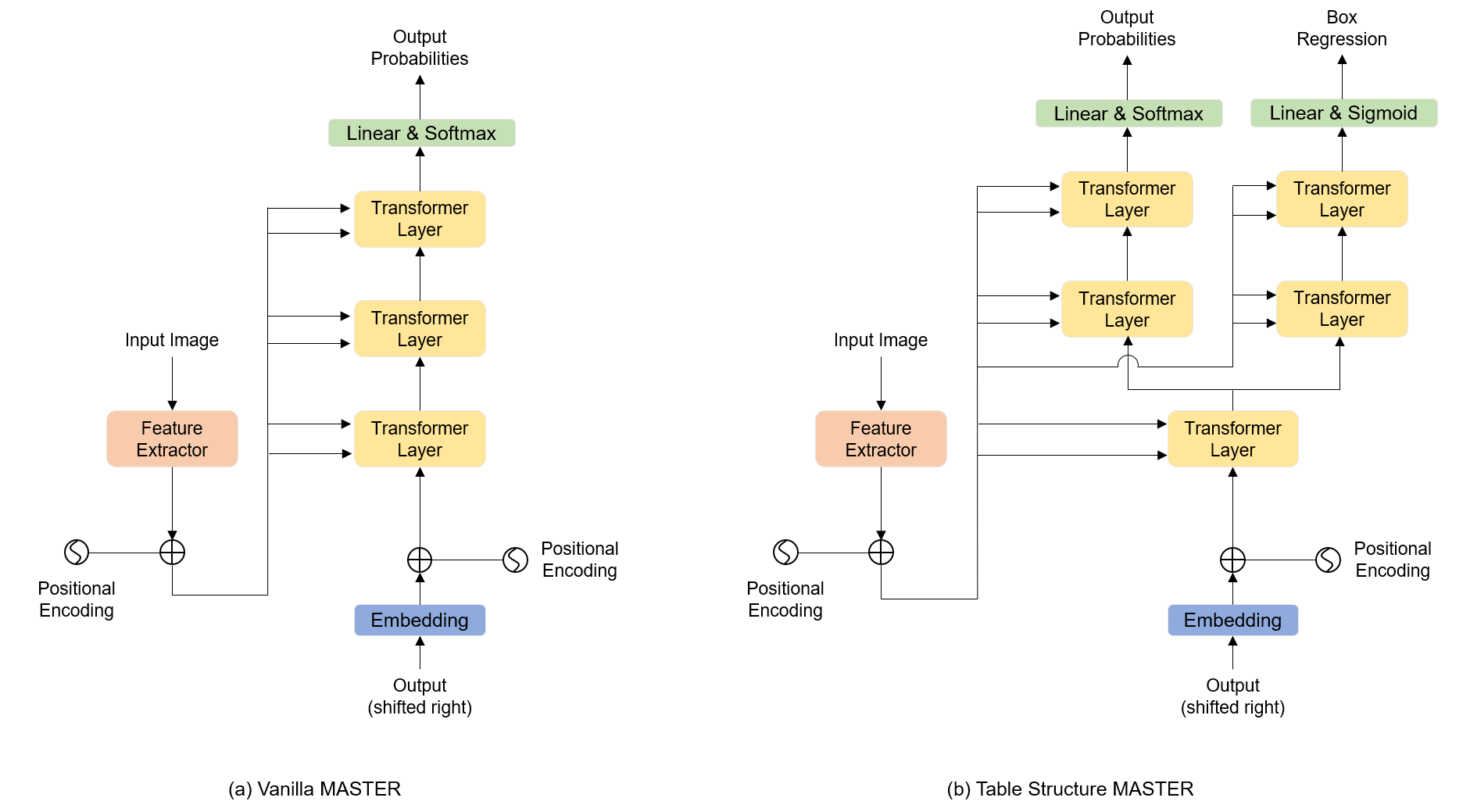}
    \caption{(a) Architecture of vanilla MASTER; (b) Architecture of table structure MASTER}
    \label{fig:structureR}
\end{figure}

\subsection{Table Structure Recognition}
The task of table structure recognition is to reconstruct the HTML sequence items and their corresponding locations on the table, but ignore the text content in each item. Our model structure is shown in Figure~\ref{fig:structureR}(b). It is customized based on MASTER~\cite{lu2019master}, a powerful image-to-sequence model originally designed for scene text recognition. Different from the vanilla MASTER as shown in Figure~\ref{fig:structureR}(a), our model has two branches. One branch is to predict the HTML item sequence, and the other is to conduct the box regression. Instead of splitting the model into two branches in the last layer,  
we decouple the 
sequence prediction and the box regression after the first transformer decode layer.

%To do this, we need to design two individual modules. As shown in  Different from the 

%Table structure recognition is a sub-task of our proposed method. In this part, we developed vanilla MASTER, a text recognition model, by adding a bounding box regression head in the decoder, which can predict structure items and their corresponding location at the same time. To decouple features, we divide two branches after feature extracting by the encoder. One branch was designed for HTML structure item prediction, another one was used for HTML cell's box regression. In vanilla MASTER, the decoder contains 3 transformer decode layers. Due to the limitation of memory, the branch-divide operation was performed after the first transformer decode layer. The architecture of vanilla MASTER was shown in Figure 1(a), and the architecture of table structure MASTER was shown in Figure 1(b).

%(b) 11 special classes and their corresponding original content.

\begin{figure}[h] % picture
    \centering
    \includegraphics[width=1\textwidth]{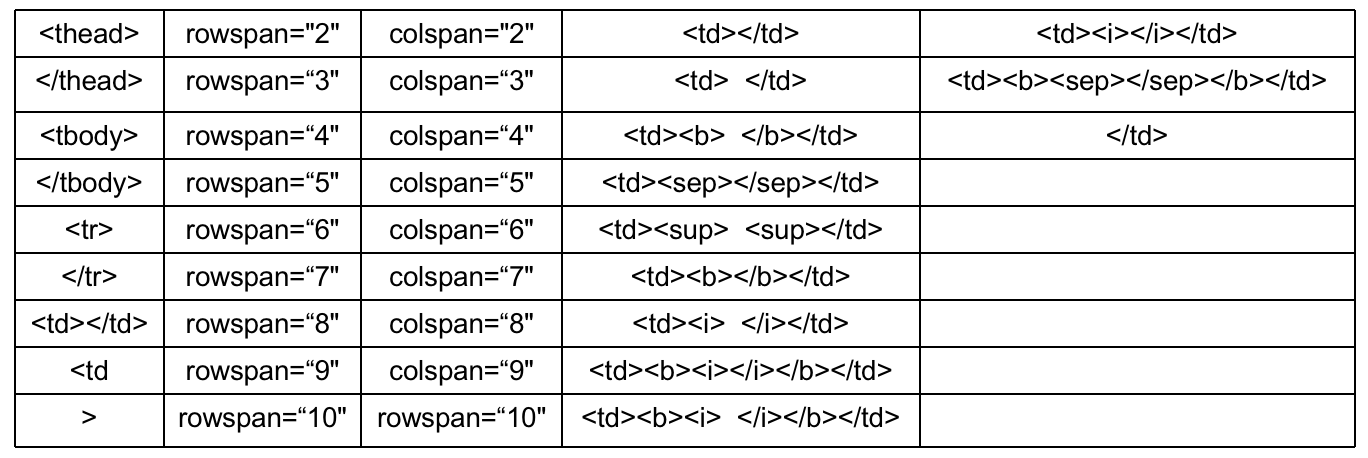}
    \caption{39 classes used in table structure MASTER. }
    \label{fig:alphabetM}
\end{figure}

To structure an HTML sequence, we need to define an Alphabet for the sequence. As shown in the left of Figure~\ref{fig:alphabetM}, we define 39 class labels for the sequence prediction. For the pairs \emph{<thead>} and \emph{</thead>}, \emph{<tbody>} and \emph{</tbody>}, and \emph{<tr>} and \emph{</tr>}, some other control characters may appear between these pairs. Thus, we need to define one individual class for each of them.
We define the maximum ``colspan'' and ``rowspan'' as 10, thus we both use 9 labels for them individually.  There are two forms for \emph{<td></td>}, empty content or non-empty content between \emph{<td>} and \emph{</td>}. 
We use one class to denote the whole of the \emph{<td>[content]</td>}. It should be noted that using one label instead of defining two individual labels for \emph{<td>} and \emph{</td>} can largely reduce the length 
of the sequence.
For the form of \emph{<td></td>} with empty content, we can find 11 special forms. As shown in the right of Figure~\ref{fig:alphabetM}, each form is represented by a special class label. According to the above description, the sequence lengths of 99.6\% HTML in the PubTabNet data set are less than 500.

%Non-empty cell position was provided by the competition dataset. It is important to note that each position labels correspond to the one non-empty \emph{<td></td>} item of table structure. In bounding box regression head, we only calculate loss at the time step which corresponding structure item is non-empty \emph{<td></td>}. 

For the sequence prediction, we use the standard cross-entropy loss. For the box regression, we employ the L1 loss to regress the coordinates of [x,y,w,h]. The coordinates are normalized to [0,1]. For the box regression head, we use an \emph{Sigmoid} activation function before the loss calculation.

\begin{figure}[h] % picture
    \centering
    \includegraphics[width=0.75\textwidth]{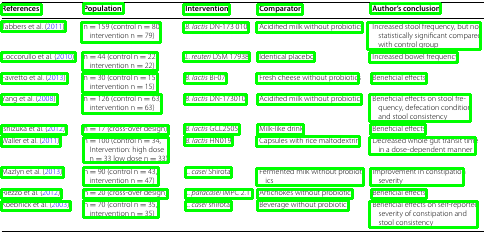}
    \caption{Example of table structure prediction. Predicted bounding box are marked with yellow color.}
    \label{fig:structureMasterExample}
\end{figure}

In Figure~\ref{fig:structureMasterExample}, we show a result example of sequence prediction and box regression. We could see that the structure MASTER can predict out the box coordinates correctly.

\subsection{Text Line Detection}
PSENet is an efficient text detection algorithm that can be considered as an instance segmentation network. It has two advantages. Firstly, PSENet, as a segmentation-based method, is able to localize texts of arbitrary shape. Secondly, the model proposes a Progressive Scale Expansion Network which can successfully identify adjacent text instances. PSENet not only adapts to text detection at arbitrary angles but also works better for adjacent text segmentation.

%PSENet is a segmentation-based text detection algorithm, which is suitable for compact text line detection. Different from box regression of table structure recognition, PSENet is used to locate every single text line in table image instead of table cell item location. These text line bounding boxes will be assigned to table cell item by box assignment algorithm, which will be discussed in detail at Section 2.4. 

\begin{figure}[h] % picture
    \centering
    \includegraphics[width=0.75\textwidth]{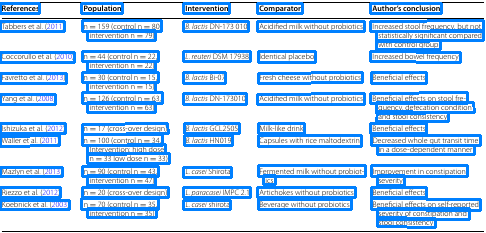}
    \caption{Visualization of text line detection.}
    \label{fig:PSENetTextLineDetection}
\end{figure}

Text detection in print documents is an easy task compared to text detection in a natural scene. In training PSENet, there are three key points needing attention, the input image size, the minimum area and the minimum kernel size.  To avoid true negative, especially some small region (such as a dash line), the resolution of the input image should be large, and the minimum area size should be set to be small. In Figure~\ref{fig:PSENetTextLineDetection},  we visualize an detection result by PSENet.

\subsection{Text Line Recognition}
We also use MASTER as our text line recognition algorithm. MASTER is powerful and can be freely adapted to different tasks according to different data forms, e.g. curved text prediction, multi-line text prediction, vertical text prediction, multilingual text prediction.

%Due to MASTER is a 2D attention-based algorithm, multi-line text images can be converted to corresponding text content.

\begin{figure}[h] % picture
    \centering
    \includegraphics[width=0.75\textwidth]{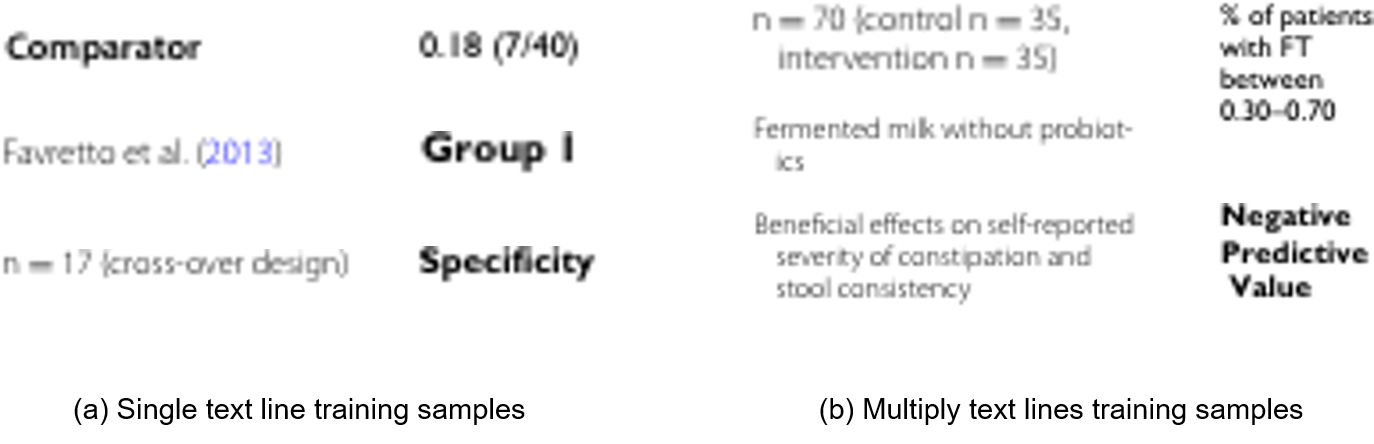}
    \caption{Example of text line images cropped from training data of PUbTabNet data set; (a) single line text image; (b) multi-lines text image}
    \label{fig:ocrtrainingsamples}
\end{figure}

The position annotations in the PubTabNet dataset (v2.0.0) is cell-level, cropped text images according to the position annotation in the data set contains both single-line and multi-line text images.
We construct a text line recognition database according to position information provided in the annotation file. This text line recognition database contains about 32 million samples cropped from 500k training images. We split out 20k text line images as a validation set for checkpoint selection. Some training samples are shown in Figure~\ref{fig:ocrtrainingsamples}. We can see that some texts are blur, and some are black and some are grey.
The maximum sequence length is set to be 100 in our MASTER OCR. Text lines longer than 100 characters will be discarded. Some training samples are shown in Figure~\ref{fig:ocrtrainingsamples}.

It should be noted that in training stage, our algorithm is trained on a database mixed with single-line text images and multi-line text images, but in the test stage, only single-line text images are inputted.
By text line recognition, we can get the corresponding text content of text line images. These text contents will be merged to non-empty \emph{<td></td>} items in the HTML sequence. The details of text content merge will also be discussed in the next subsection.

\subsection{Box Assignment}
According to the above three subsections, we have obtained the table structure together with the box of each cell, and the box of each text line together with its corresponding text content. To generate the complete HTML sequence, we need to assign each box of text line into its corresponding table structure cell.
In this subsection, we will introduce our used match rules in detail. There are three matching rules used in our method, which we call \emph{Center Point Rule}, \emph{IOU Rule} and \emph{Distance Rule}. The details will be discussed below.

\subsubsection{Center Point Rule}
In this matching rule, we firstly calculate central coordinate of each box obtained by PSENet. If the coordinate is in the rectangular region of the regressed box obtained by structure prediction, we call them a matching pair. The content of the text line will be filled into \emph{<td></td>}.
It is important to note that one table structure cell can be associated with several PSENet boxes because of one table structure cell may have multiple text lines.

\subsubsection{IOU Rule}
If the above rule is not satisfied, we will compute the IOU between the box of the chosen text line and all structure cell boxes. The box cell with the maximum IOU value will be selected. The text content will be filled into the chosen structure cell.
%For one PSE box, we will calculate their IOU with each MASTER boxes. The MASTER box with the maximum IOU value and the maximum value greater than IOU threshold, will be attached to corresponding PSE box.

\subsubsection{Distance Rule}
Finally, if both above rules are unsuccessful. We will calculate the Euclidean distances between between the box of the chosen text line and all structure cell boxes. Similar to the \emph{IOU Rule}, the structure cell with minimum Euclidean distance will be chosen.

\subsubsection{Matching Pipeline}
All above-mentioned three rules will be applied in order. Firstly, most boxes detected by PSENet will be assigned to their corresponding structure cells by \emph{center point rule}. Owing to prediction deviations of structure prediction, a few central points of PSENet boxes are out of the rectangle region of structure cell boxes obtained by structure prediction. Secondly, some unmatched PSENet boxes under the \emph{center point rule} will get matched under the \emph{IOU Rule}. In the above two steps, we use the PSENet boxes to match their corresponding structure item. If there are some structure items that are not matched. In this way, we use the structure item to find the left PSENet boxes. To do this, the \emph{distance rule} is applied.

\begin{figure}[h] % picture
    \centering
    \includegraphics[width=1.0\textwidth]{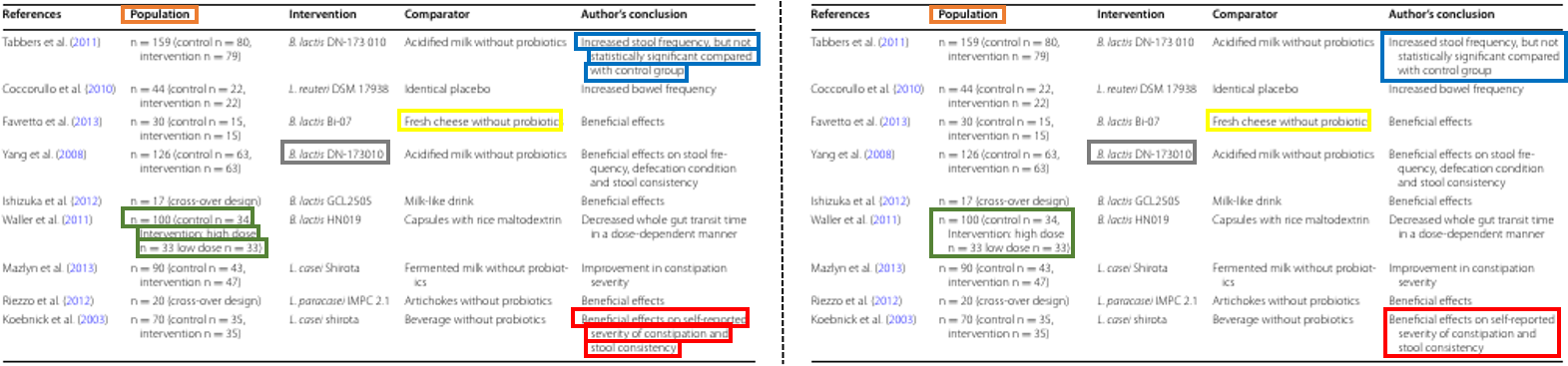}
    \caption{Example of box assignment visualization. On the left side, some detected boxes by PSENet are marked by different colors. On the right side, the boxes generated by structure prediction are marked.}
    \label{fig:matchingresults}
\end{figure}

A visualization example of matching results is shown in Figure~\ref{fig:matchingresults}. For aesthetic effect, we only show part of the boxes.
On the left side of Figure~\ref{fig:matchingresults}, some detected boxes by PSENet are marked by different colors. On the right side of Figure~\ref{fig:matchingresults}, the boxes generated by structure prediction are marked. The boxes on the left side will be assigned to the box cell with the same color. 

%is prediction of PSENet, which detection every single text line of images. Box regression result of table structure MASTER was shown in the right side. The number at the upper left corner is matching pairs index. The same number of PSENet prediction and table structure MASTER box prediction will be regarded as a matching pair. 

\section{Experiment}
\label{sec:others}
In this section, we will describe the implementation of our table recognition system in detail.

{\bf{Dataset.}} Our used data is the PubTabNet dataset (v2.0.0), which contains 500,777 training data and 9,115 validation data in development Phase, 9,138 samples for the development stage, and 9,064 samples for the final evaluation stage.
Except for the provide training data, no extra data is used for training. To get text-line level annotation of all text boxes, 2k images of training data are relabeled for PSENet training. Actually, we only need to adjust the annotations of multi-line annotation into single-line box annotation.

%For once end-to-end inference, one table image will forward PSENet, table structure MASTER and text line recognition MASTER. Box assignment(matching rules) and some HTML format rules will be used to get the final HTML code.

{\bf{Implementation Details.}} In PSENet training, 8 Tesla V100 GPUs are used with the batch size 10 in each GPU. The input image is resized equally, keeping the long side with resolution 1280. RandomFilp and RandomCrop are used for data augmentation. A $640\times 640$ region is cropped from each image. 
Adam optimizer is applied, and the initial learning rate is 0.001 with step learning rate decay.

In table structure training, 8 Tesla V100 GPUs are used with the batch size 6 in each GPU. The input image size is $480\times 480$, and the maximum sequence length is 500. Synchronized BN~\cite{zhang2018context} and Ranger optimizer~\cite{lessw2019ranger} are apply in this experiment, and the initial learning rate of optimizer is 0.001 with step learning rate decay.

In the training of text line recognition, 8 Tesla V100 GPUs are used with the batch size 64 in each GPU. The input size is $256\times 48$, and the maximum length is 100. Synchronized BN and Ranger optimizer are also applied and the hyper-parameter setting is the same as the table structure training. 

All models are trained based on our own FastOCR toolbox.

\subsection{Ablation Studies}
Our table recognition system is described above. We have conducted many attempts in this competition. In this subsection, we will discuss some useful tricks, but ignore some unsuccessful attempts.

{\bf{Ranger}} is a synergistic optimizer combining RAdam (Rectified Adam)~\cite{liu2019radam}, LookAhead~\cite{zhang2019lookahead}, and GC (gradient centralization)~\cite{yong2020gradient}. We observe  that Ranger optimizer shows a better performance than Adam in this competition, and it is applied in both table structure prediction and text line recognition. We use default Ranger.
Result comparison between Adam and Ranger is shown in Table~\ref{tab:ablationstudy}(a).

{\bf{Synchronized Batch Normalization (SyncBN) }} is an effective batch normalization approach that is suitable for multi-GPU or distributed training. In standard batch normalization, the data is only normalized within the data on each GPU device. But SyncBN normalizes the input within the whole mini-batch. SyncBN is ideal for situations where the batch size is relatively small on each GPU graphics card.
SyncBN is applied in our experiment.

{\bf{Feature Concatenation of Layers in Transformer Decoder.}}
In structure MASTER and text recognition MASTER, three successive transformer layers~\cite{lu2019master} is used as decoder. Different from the original MASTER, we concatenate the outputs of each transformer layer~\cite{dou2018exploiting} and then apply a linear projection on the concatenated feature.

{\bf{Label Encoding in Structure Prediction}}
After we inspect on the training data of the PubtabNet data set(v2.0.0), we find some ambiguous annotations about empty table cell. Some empty cells of table are labeled as \emph{<td></td>}, whereas the others are labeled as \emph{<td> </td>} in which one space character is inserted. However, these two different table cells look the same visually. According to statistics, the ratio between \emph{<td></td>} and \emph{<td> </td>} is around 4:1. In our experiment, we encode these two different cells into different tokens. Our motivation is to let the model to discover the intrinsic visual features by training. 
%Encoding \emph{<td></td>} and \emph{<td> </td>} separately, it improves 0.8 TEDS score. It's important to note that, taking this label encoding method for training, word accuracy will decrease. However, it performance greater in final TEDS evaluations. 

%added by QXB

\begin{table*}[htbp]
  \centering
  \subfloat[Comparison of optimizer. \label{tab:chapter5:1a}]{
    \normalsize
    \centering
 \begin{tabular}{l|c}
\hline
\textbf{Optimizer}                & \multicolumn{1}{l}{\textbf{Structure prediction Acc.}} \\ \hline
Adam & 0.7784                                            \\ \hline
Ranger         & \textbf{0.7826}                                            \\ \hline
\end{tabular}
  }
  \hspace{10pt}
  \subfloat[Comparison of with or without feature concatenation.\label{tab:chapter5:1b}]{
    \normalsize
    \centering
     \begin{tabular}{c|c}
\hline
\textbf{Feature Concatenation}            & \textbf{Text line recognition Acc.} \\ \hline
No & 0.9313                               \\ \hline
Yes        & \textbf{0.9347}                               \\ \hline
\end{tabular}
  }

  \subfloat[Evaluation of label encoding, SyncBN and feature concatenation. \label{tab:chapter4:1f}]{
    \normalsize
    \centering
    \begin{tabular}{cc|c}
\hline
\textbf{SyncBN} & \textbf{FC} & \textbf{Structure prediction Acc.} \\ \hline
                 &             & 0.7734                        \\ \hline
\checkmark               &             & 0.7750                        \\ \hline
\checkmark               & \checkmark           & \textbf{0.7785}                        \\ \hline
\end{tabular}
  }
  \caption{Evaluation of different tricks on table recognition task. (a). comparison of Ranger and Adam. (b). comparison of with or without feature concatenation. (c). evaluation of label encoding.}\label{tab:chapter4:1}
\label{tab:ablationstudy}
\end{table*}

In this competition, we have conducted some evaluations and recorded the results. The results are shown in Table~\ref{tab:ablationstudy}.

According to Table 1, we have the following observations,
\begin{itemize}[leftmargin=.1in]
  \item Ranger optimizer has outperformed Adam optimizer consistently. Similar observation is also found in our another report~\cite{he2021ICDAR} about ICDAR 2021 Competition on Scientific Table Image Recognition to LaTeX~\cite{Pratik2021ICDAR}. In our evaluation on standard benchmarks, we also find that Ranger can improve the average accuracy by around 1\%. 
  \item SyncBN can improve the performance a little. We also observe that SyncBN also shows better performance than standard BN on ICDAR 2021 competition on Mathematical Formula Detection.
  \item Feature concatenation can improve the accuracy of the structure prediction on this task. It should be noted that in~\cite{he2021ICDAR}, we do not observe performance improvement.
\end{itemize}

\begin{table}[h]
\centering
\begin{tabular}{|c|c|c|c|c|c|c|c|c|}
\hline
\textbf{TLD} & \multicolumn{3}{c|}{\textbf{TSR}} & \textbf{TLR} & \textbf{BA}  & \multirow{2}{*}{\textbf{ME}} & \multirow{2}{*}{\textbf{ForC}} & \multirow{2}{*}{\textbf{TEDS}} \\ \cline{1-6}
PSE          & ESB       & SyncBN      & FeaC      & FeaC           & Extra Insert &                              &                                 &                                           \\ \hline
\checkmark            &           &             &         &              &              &                              &                                 & 0.9385                                    \\ \hline
\checkmark            & \checkmark         & \checkmark           & \checkmark       &              & \checkmark            &                              &                                 & 0.9621                                    \\ \hline
\checkmark            & \checkmark         & \checkmark           & \checkmark       & \checkmark            & \checkmark            &                              &                                 & 0.9626                                    \\ \hline
\checkmark            & \checkmark         & \checkmark           & \checkmark       & \checkmark            & \checkmark            & \checkmark                            &                                 & 0.9635                                    \\ \hline
\checkmark            & \checkmark         & \checkmark           & \checkmark       & \checkmark            & \checkmark            & \checkmark                            & \checkmark                               & \textbf{0.9684}                                    \\ \hline
\end{tabular}
\caption{End-to-end evaluation on the validation set with TEDS as the indicator. TLD: text line detection; TSR: table structure recognition; TLR: text line recognition; ME: model ensemble. ESB: empty space box encode; SyncBN: synchronized BN; FeaC: feature concatenate output of transformer layers. ForC: format correction.}
\label{tab:endtoendEval}
\end{table}

\subsection{End-to-end Evaluation on the Validation Set}
We generate the final HTML code by merging structure prediction, text line detection, text line recognition, and box assignment. We evaluate some tricks in these stages. Results are shown in Table~\ref{tab:endtoendEval}. TEDS is used as our indicator.

%The end-to-end result was shown in Table 4. Noting that \emph{Extra Insert} in Table 4, means assign no matching PSE boxes and corresponding table contents to the end of HTML code. By using this trick in our experiment, TEDS score will be improved.    

We have some overall conclusions from this competition,
\begin{itemize}[leftmargin=.1in]
\item ESB (empty space box encode) is important for the final TEDS indicator.
\item FeaC (feature concatenation) is effective for both table structure recognition and text line recognition.
\item ME (model ensemble) improves the performance a little bit. Three model ensembles in the TSR can improve the end-to-end TEDS score for around 0.2\%. Three model ensembles in the text line recognition can only improve the TEDS score for around 0.03\%. We only use one PSENet model.
\item SyncBN is effective for both TSR and TLR.
\item ForC (format correction) helps the final indicator. Our format correction is to promise all content between \emph{<thead>} and \emph{</thead>} is black font.
\end{itemize}

\begin{figure}[h] % picture
    \centering
    \includegraphics[width=0.90\textwidth]{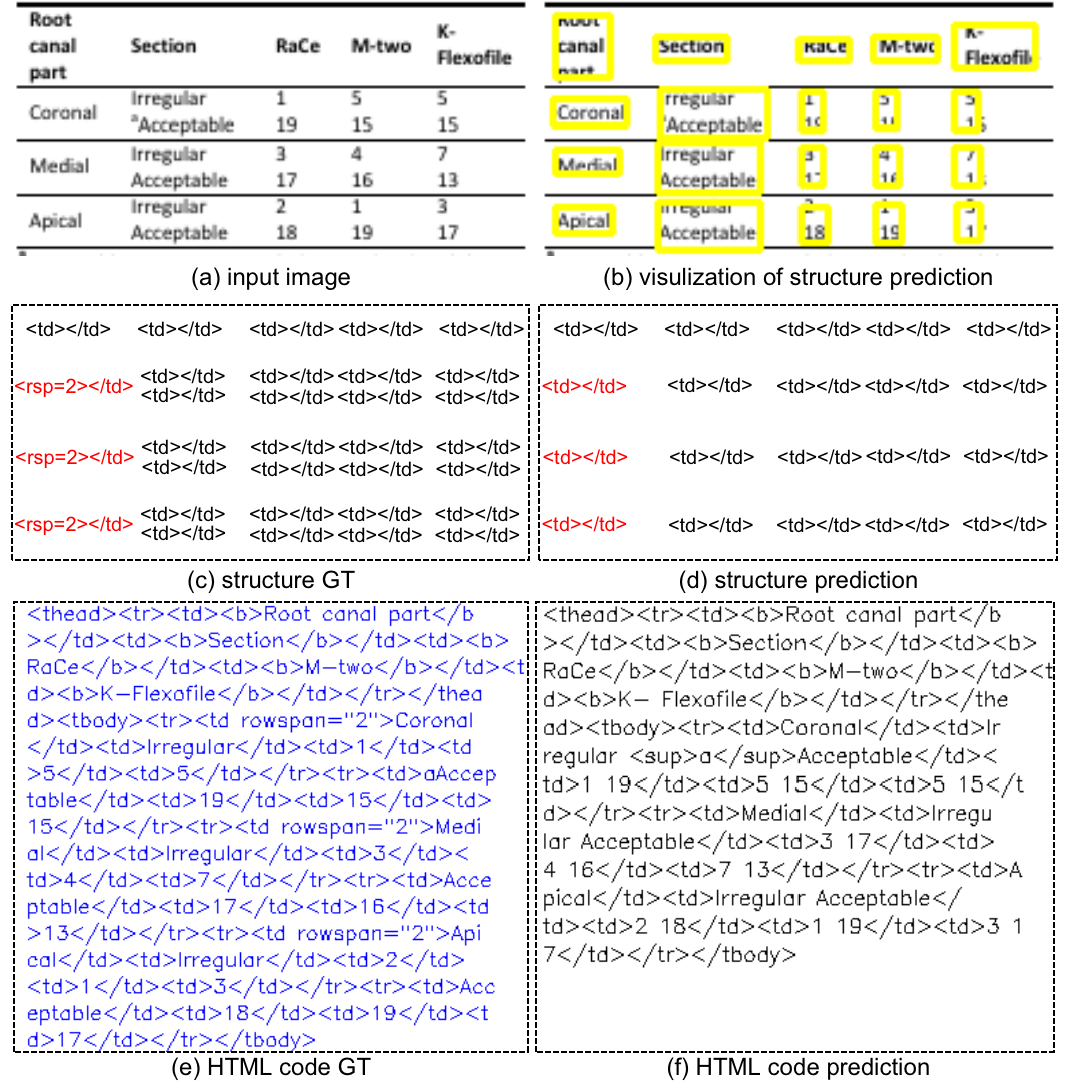}
    \caption{An example of wrong table structure prediction.}
    \label{fig:wrongexample}
\end{figure}

{\bf{Discussion.}} From this competition, we have some reflections. For the end-to-end table recognition to the HTML code, structure prediction is an extremely important stage, especially for the TEDS indicator. As shown in Figure~\ref{fig:wrongexample}, although all text line information is correctly recognized. Our method obtains very low TEDS (0.423) due to wrong structure prediction.
Although the provided data set is large, we still believe larger scale of data that cover more templates may further improve the structure prediction.
Secondly, text line detection and text line recognition are easy tasks considering all table images are print. Thirdly, There are some labeling inconsistency issues, such as 
\emph{<td></td>} and \emph{<td> </td>}. Finally, 
the box assignment sub-task can be conducted by Graph Neural Network (GNN)~\cite{chen2020learning} instead of hand-crafted rules.

\section{Conclusion}
\label{sec:conclusion}
In this paper, we present our solution for the ICDAR 2021 competition on Scientific Literature Parsing task B: table recognition to HTML. We divide the table recognition system into four sub-tasks, table structure prediction, text line detection, text line recognition, and box assignment. Our system gets a 96.84 TEDS scores on the validation data set in the development phase, and gets a 96.324 TEDS score in the final evaluation phase.

\bibliographystyle{unsrt}  
\bibliography{references}

\end{document}